\documentclass[10pt,twocolumn,letterpaper]{article}
\pdfoutput=1

\usepackage{iccv}
\usepackage{times}
\usepackage{epsfig}
\usepackage{graphicx}
\usepackage{amsmath}
\usepackage{amssymb}
\usepackage{booktabs}
\usepackage{multirow}
\usepackage{multicol}
\usepackage{amssymb}
\usepackage{bm}

\newcommand{\mytilde}{\raise.17ex\hbox{$\scriptstyle\mathtt{\sim}$}}
\DeclareMathOperator*{\argmin}{arg\,min}

\usepackage[pagebackref=true,breaklinks=true,letterpaper=true,colorlinks,bookmarks=false]{hyperref}

\iccvfinalcopy

\def\iccvPaperID{3887}

\ificcvfinal\fi

\begin{document}

\title{\textit{SurfaceNet}: Adversarial SVBRDF Estimation from a Single Image}

\author{
Giuseppe Vecchio \qquad Simone Palazzo \qquad Concetto Spampinato\\
PeRCeiVe Lab\\
Department of Electical, Electronic and Computer Engineering -- University of Catania\\
{\tt\small www.perceivelab.com}
}

\maketitle
\ificcvfinal\thispagestyle{empty}\fi

\begin{abstract}
In this paper we present \textbf{SurfaceNet}, an approach for estimating spatially-varying bidirectional reflectance distribution function (SVBRDF) material properties from a single image.
We pose the problem as an image translation task and propose a novel patch-based generative adversarial network (GAN) that is able to produce high-quality, high-resolution surface reflectance maps. The employment of the GAN paradigm has a twofold objective: 1) allowing the model to recover finer details than standard translation models; 2) reducing the domain shift between synthetic and real data distributions in an unsupervised way. \\ 
An extensive evaluation, carried out on a public benchmark of synthetic and real images under different illumination conditions, shows that \textit{SurfaceNet} largely outperforms existing SVBRDF reconstruction methods, both quantitatively and qualitatively.
Furthermore, \textit{SurfaceNet} exhibits a remarkable ability in generating high-quality maps from real samples without any supervision at training time. 

Source code available at {\tt\small \url{https://github.com/perceivelab/surfacenet}}.

\end{abstract}

\section{Introduction}
Computer-generated imagery (CGI) and 3D graphics play an important role in a variety of applications, including visual effects, architectural modeling, simulators, cultural heritage, video games, virtual or augmented reality and automotive design. 
The increasing computational power of both professional and consumer hardware has led to a growing interest in high-quality CGI, whose basic requirement for creating realistic images is the definition and implementation of robust digital models that describe how real-world materials interact with light~\cite{dorsey2005digital,10.5555/1557600}.
This task is easily carried out by humans, who are able to intuitively identify materials' physical properties by analyzing how light is reflected, transmitted and absorbed before reaching the observer's eyes. Artificially emulating this process would require a physically-accurate simulation of how generic materials interact with light, but the complexity of such task and the needed level of surface details make this approach computationally unfeasible. 

In practice, most approximations for rendering reflections over surfaces simplify the task by defining a model that describes how light interacts with pixel-level elements of a material depicted in a digital image: the properties of the material are modeled by a \emph{spatially-varying bidirectional reflectance distribution function} (SVBRDF) that is parameterized by a set of properties encoding color, planar deformation and reflectivity.
However, even measuring this approximation is a major challenge in computer graphics. 

Following the success of deep learning methods in computer vision, the estimation of material reflectance properties has been increasingly posed as a learning task~\cite{deschaintre2018single,gao2019deep,deschaintre2019flexible,deschaintre2020guided,guo2020materialgan}. Following this trend, we propose \emph{SurfaceNet}, a fully-convolutional network for SVBRDF estimation from a single input image. Unlike methods that estimate material properties from multiple input images~\cite{deschaintre2019flexible,gao2019deep}, our approach better suits non-professional application scenarios, where it is unfeasible to obtain reliable acquisitions of a surface with a sufficiently steady view point.

More specifically, we pose SVBRDF estimation as an \emph{image-to-image translation} task and introduce a deep generative adversarial architecture, consisting of a \textit{generator} that employs a fully-convolutional multi-head encoder-decoder network that predicts a set of SVBRDF maps, and a patch-based \textit{discriminator} that is trained to distinguish between estimated maps and ground-truth ones. 
We propose to use a generative adversarial loss to compensate for the blurriness typically introduced by $L_1$ or $L_2$ losses. This is an alternative approach to recent methods that perform \emph{neural rendering}~\cite{deschaintre2018single,gao2019deep} to produce output images from the estimated reflectance maps as a supervisory signal. 
Moreover, the GAN framework also allows us to employ real-world images, for which no reflectance maps are available, during the training procedure, alongside synthetic images. As a result, our generator learns to extract features that can be shared between synthetic and real images, enforcing an implicit domain adaptation mechanism and reducing the distribution gap between input images from the two modalities. 
the combined use of skip connections within the generator and of a patch-based discriminator allows the model to focus on and recover detailed features of small patches. We argue that these characteristics are particularly appropriate for SVBRDF estimation of real-world materials, where the capability to work at high-resolution on surface details is essential and where pattern structure is generally local.

We evaluate our method on a wide variety of materials from publicly-available SVBRDF libraries and real-world pictures of surfaces; in our experiments on single-image inputs, our method largely outperforms previous works, both qualitatively and quantitatively.

To summarize, the contributions introduced by the proposed method are the following: 
\begin{itemize}
\item We present a deep network for single-image SVBRDF estimation that, leveraging the properties of GANs in generating high-frequency details and learning data distributions in an unsupervised way, is able to predict high-quality reflectance maps from real-world photographs; 
allows our model to recover fine, local, features in the output maps, even at high resolution (2048$\times$2048); 
\item Experimental results on multiple datasets under different illumination conditions show that our model largely outperforms, both quantitatively and qualitatively, existing single-image estimation methods, setting new state-of-the-art performance on the task.
\end{itemize}

\section{Related work}
\begin{figure*}
\centering
\includegraphics[width=1\linewidth]{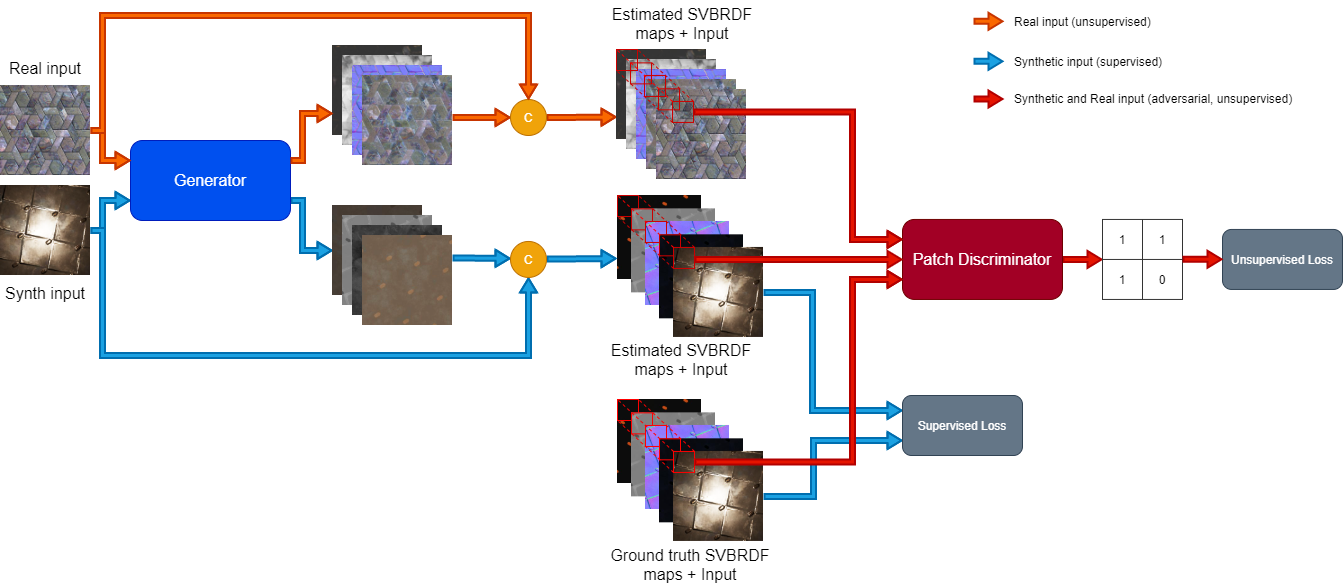}
\caption{\textbf{Overview of the SurfaceNet framework}. An input image is fed to the generator, which estimates SVBRDF parameter maps. A discriminator receives patches of SVBRDF maps and attempts to distinguish between estimated maps (from both real and synthetic images) and ground-truth maps (for synthetic images only). A supervised loss term (based on $L_1$ norm and MS-SSIM similarity) is computed on the output maps from the generator using ground-truth SVBRDF maps. An adversarial unsupervised loss term is instead computed for the patch discriminator. Circled ``C'' blocks indicate feature concatenation.}
\label{fig:overview}
\end{figure*}

Surface reflectance models describe how light reflects on opaque surface points by means of mathematical equations that are a function of the directions of incidence and outgoing radiance. Approximations of this phenomenon for non-uniform materials are defined by \emph{spatially-varying bidirectional reflectance distribution functions} (SVBRDF), which are typically parameterized by a set of 2D maps that encode the specific properties of a material or a surface, e.g. diffuse albedo, specular albedo, normal, roughness, ambient occlusion~\cite{dong2019deep}.

Traditionally, surface reflectance models are tied to specific reconstruction algorithms, which are able to estimate reflectance properties for certain materials only, by employing hand-crafted heuristics based on distribution priors~\cite{aittala2013practical,lombardi2012single,chen2014reflectance,zhou2016sparse}. This limits the applicability of these techniques to more general contexts.

Recent data-driven approaches based on deep learning have been applied to learn such heuristics automatically from data, by providing input images and having the model extract surface reflectance properties. We hereby focus on single-image estimators, although multi-view approaches have also been proposed in the literature (e.g.,~\cite{aittala2016reflectance,deschaintre2019flexible,bi2020deep}). In~\cite{li2017modeling}, a U-net architecture is employed to extract spatially-varying diffuse albedo and normal maps, while a simpler convolutional architecture estimates homogeneous specular albedo and roughness. The approach in~\cite{li2018materials} also estimates spatially-varying roughness, by assuming the presence of a dominant flashlight, through a multi-decoder convolutional architecture and a material classifier, followed by CRF optimization for refinement. In~\cite{deschaintre2018single}, the authors introduce a \emph{rendering loss function} that decouples training from the specific parameterization of the employed surface reflectance model, and directly supervises the learning process by rendering the image through the estimated parameters and comparing it with the target. An extension of this idea is presented in~\cite{li2018learning}, where a cascaded estimation and rendering architecture is employed to better model global illumination. A deep inverse rendering model is proposed in~\cite{gao2019deep}, where an auto-encoder on the surface reflectance maps (initialized through a state-of-the-art model and estimated through neural rendering) is trained to learn a latent representation, rather than SVBRDF parameters directly.

The key difference between existing works and ours lies in how fine details are reconstructed: \cite{deschaintre2018single,li2018learning,gao2019deep} integrate a neural rendering method for reconstructing detailed features; we, instead, use the representational capabilities of GANs to learn high-frequency local visual components, complementing the global features learned by optimizing a stamdard $L_1$ loss. 
Furthermore, adversarial training enables unsupervised learning from unannotated real images, by encouraging our generator to handle the representational shift between synthetic images (used for supervised training) and real-world data. 

Guo et al. recently proposed a generative method,  MaterialGAN~\cite{guo2020materialgan}, that is trained to generate plausible materials from a learned latent space and uses the trained generator as a prior for SVBRDF estimation by iteratively optimizing a rendering loss. While this approach shares the adversarial training paradigm with ours, we directly estimate reflectance maps from a surface picture. Moreover, we make full use of the potential of the GAN paradigm, by including unannotated real images into the training loop, and demonstrate their positive impact on reconstruction accuracy.

\section{Method}
The objective of our work is to estimate the pixel-level reflectance properties of a spatially-varying material from a single input image. 
We assume that the considered surface is mostly planar and that non-planar surface details can be modeled by a normal map. In the implementation that we present here, we approximate surface reflectance at each point through the Cook-Torrance model using the GGX microfacet distribution~\cite{walter2007microfacet}. However, our approach can be indifferently applied to any reflectance model whose properties can be estimated in terms of spatial maps.

It is well known that, in tasks where the supervisory signal consists in whole images (e.g., image synthesis, image-to-image translation, or the task at hand), $L_1$ and $L_2$ reconstruction losses are able to enforce correctness at the low frequencies, but tend to produce blurry results and miss high-frequency details~\cite{isola2017image,larsen2016autoencoding}. We overcome this limitation by complementing a reconstruction loss with an adversarial loss, and training a discriminator network to distinguish whether an input set of material maps is produced by the generator or sampled from the training set. At the same time, the generator is also trained to maximize the probability of the discriminator believing that the estimated maps have the same quality as the ground truth. As a result, both models simultaneously improve, and in particular the generator is pushed --- beyond the limits of the $L_1$ loss --- to produce output maps that are as realistic as possible.
Furthermore, we train the whole model with two sets of data: a set of annotated synthetic images to supervisedly enhance the overall estimation quality, and a set of real images without annotations to allow the model to learn, in an unsupervised way, how to correctly estimate reflectance maps in case of real-world input images.

We design our method, \emph{SurfaceNet}, as an image-to-image translation problem within a generative adversarial framework, where the image of a planar material is translated into the corresponding set of SVBRDF maps. An overview of our approach is shown in Fig.~\ref{fig:overview}. The generator network receives an input image and acts as our SVBRDF estimator, by providing surface reflectance maps as output. These maps are then fed to the discriminator network, which aims at distinguishing them from ground-truth maps from the training set. 
During training, the generator and discriminator adversarially compete, with the former trying to mislead the latter by generating more and more realistic maps, and the latter learning to identify which maps are produced by the generator.

In the following, we introduce and describe each module of the proposed framework. Architectural details of individual layers are included in the supplementary materials.

\begin{figure}
\centering
\includegraphics[width=0.8\linewidth]{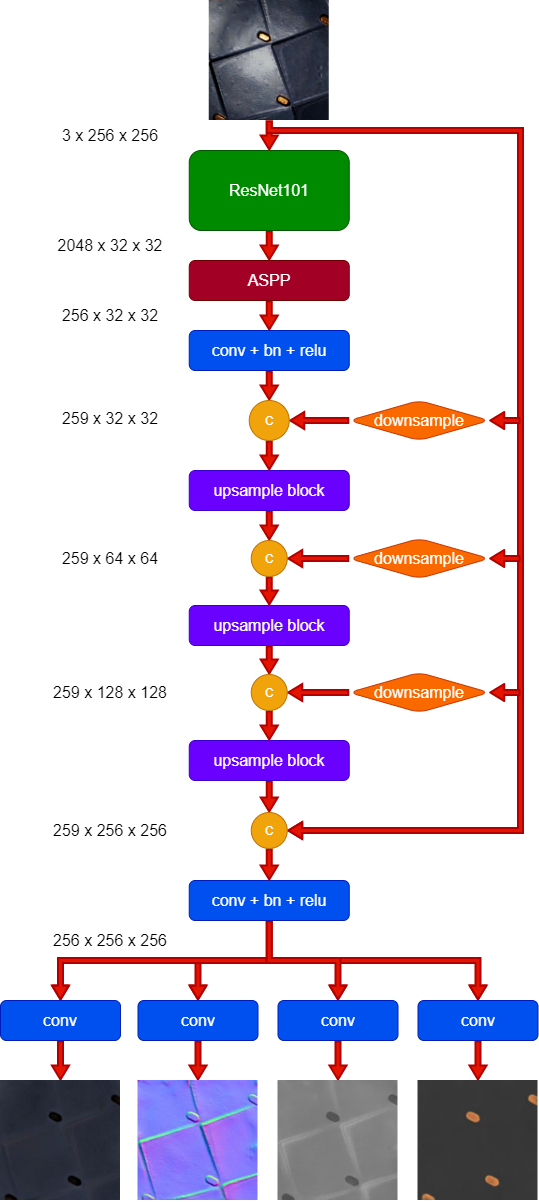}
\caption{Overview of the architecture of the SurfaceNet generator. The number of heads is variable and depends on the number of maps to predict.}
\label{fig:generator}
\end{figure}

\subsection{Generator architecture}
The generator network in SurfaceNet, illustrated in Fig.~\ref{fig:generator}, is inspired by the architecture of DeepLabV3~\cite{chen2017rethinking}, an encoder-decoder semantic segmentation model based on ResNet-101~\cite{he2016deep}. The input to the model is an RGB image of arbitrary size, since the architecture is fully-convolutional; in our experiments, we train our model on images of size 256$\times$256. The encoder of our generator consists of a variant of ResNet-101 followed by Atrous Spatial Pyramid Pooling (ASPP)~\cite{chen2017deeplab} to extract multi-scale features. The output of the DeepLabV3 model has size 32$\times$32: in order to recover the original size of the image, we append a cascade of upsampling blocks, implemented as transposed convolutions alternated to residual layers.  Each upsampling block also receives a correspondingly downsampled copy of the input through skip connections, in order to provide information useful to reconstruct fine details. After the upsampling stage, the model produces a set of 256 feature maps, each of size 256$\times$256. The final reflectance maps are obtained by feeding these shared feature maps to independent prediction heads. Note that most of the computation carried out by the model is shared by all output maps, thus improving efficiency, encouraging feature reuse and allowing the model to correlate information across different reflectance parameters.

\subsection{Patch discriminator architecture}
\label{subsec:patch-discr}
The architecture of the discriminator is inspired from the original work of Isola et al.~\cite{isola2017image}. The network consists of 6 convolutional layers, such that the spatial size of the output feature maps is reduced by a factor of 18. As a consequence, a set of 256$\times$256 maps is reduced to a 1-channel map of size 14$\times$14. We treat this output map as a set of patch-level scalar predictions by the discriminator. This allows the discriminator to work independently on overlapping patches, returning a prediction for each patch, and to focus on the reconstruction of local details. The set of responses for each patch is then averaged to provide the final output of the discriminator. The patch-based discriminator is particularly suitable for material surface reconstruction, which requires identifying and recovering fine details but where global structure is generally lacking, and can be recovered by associating a standard $L_1$ loss.
The architecture of the patch discriminator is described in detail in the supplementary materials.

\subsection{Training strategy}
\label{sec:training}

Formally, given input image $\textbf{I}$, representing a real surface of a certain material or the rendering of a synthetic image, and the corresponding reflectance maps $\left\{ \textbf{M}_1, \textbf{M}_2, \dots, \textbf{M}_k\right\}$ (with $k$ depending on the employed surface reflectance model), and given a neural network $G$ (i.e., our generator) that estimates a set of approximated reflectance maps $\left\{ \hat{\textbf{M}}_1, \hat{\textbf{M}}_2, \dots, \hat{\textbf{M}}_k\right\}$ from $\textbf{I}$, the objective of the training procedure is to optimize the parameters $\bm{\theta}$ of the neural model $G$ and minimize a loss function encoding the approximation error:
\begin{equation}
\argmin_\theta \sum_i \mathcal{L}\left( \textbf{I}_i, \textbf{M}_{i,1}, \dots, \textbf{M}_{i,k} \right)
\end{equation}
with $i$ iterating over the training dataset. \\
The training strategy includes two different streams: one, \emph{supervised}, applied when feeding the model with synthetic data and corresponding ground-truth maps, and another, \emph{unsupervised}, where instead we feed real data to the model and do not use any annotations. Consequently, the overall loss $\mathcal{L}$ consists of two terms --- a supervised loss (which acts as a reconstruction loss) and an adversarial unsupervised loss:
\begin{equation}
\mathcal{L} = \mathcal{L}_\text{sup} + \alpha \mathcal{L}_\text{unsup}
\end{equation}
weighed by a hyperparameter $\alpha$.

The supervised loss $\mathcal{L}_\text{sup}$ is computed only on images for which SVBRDF maps are available at training time, and consists in a reconstruction loss that  evaluates the global similarity between ground-truth maps and the maps predicted by the generator. $\mathcal{L}_\text{sup}$ is specifically composed by: 1) a $L_1$ loss term that compares each pixel independently, and 2) an MS-SSIM~\cite{wang2003multiscale} loss term that preserves high-frequency contrast but is not sensitive to uniform biases, possibly causing changes of brightness or colors~\cite{zhao2015loss}. Therefore, the supervised loss is computed as follows:

\begin{equation}
\mathcal{L}_\text{rec} = \sum_k \left[ \left\lVert \textbf{M}_k - \hat{\textbf{M}}_k \right\rVert_1 + \beta~\text{MS-SSIM}\left(\textbf{M}_k, \hat{\textbf{M}}_k \right) \right]
\end{equation}
where $k$ iterates over reflectance maps, $\text{MS-SSIM}(\cdot,\cdot)$ computes the MS-SSIM similarity between a pair of maps, $\beta$ acts as a weighing factor. and $\left\{ \hat{\textbf{M}}_1, \dots, \hat{\textbf{M}}_k \right\} = G(\mathbf{I})$.

As unsupervised loss $\mathcal{L}_\text{adv}$, we use the standard GAN adversarial loss at patch level, that aims at pushing the predictor $G$ to synthesize patches that are indistinguishable from ground-truth ones to a discriminator $D$, while training the same discriminator to improve at separating the two data sources.
Assuming that the output $D\left( \textbf{M}_1, \dots, \textbf{M}_k \right)$ of the discriminator is a scalar likelihood value, computed as the mean of patch-level predictions, we can define the adversarial loss $\mathcal{L}_\text{sup}$ as the sum of a discriminator loss $\mathcal{L}_\text{disc}$ and a generator loss $\mathcal{L}_\text{gen}$, that can be respectively computed for a single sample as follows:
\begin{equation}
\mathcal{L}_\text{disc}
= \log D\left( \textbf{M}_1, \dots, \textbf{M}_k \right)
+ \log \left( 1 - D\left( G\left( \textbf{I} \right) \right) \right)
\end{equation}

\begin{equation}
\mathcal{L}_\text{gen}
= \log D\left( G\left( \mathbf{I} \right) \right)
\end{equation}

We apply the adversarial loss both on synthetic data with ground truth and on unannotated real data. As a result, the generator improves at estimating reflectance maps with high-frequency details, while at the same time filling the domain gap between synthetic and real images, by learning input feature representations that are equally applicable to both data sources.

\section{Experimental results}
\label{sec:experiments}

In this section, we first introduce the datasets employed in our work: the synthetic dataset with SVBRDF annotation, and the real dataset that we employ for unsupervised domain adaptation.

Then, we evaluate the accuracy of our approach on two different training setups. First, we assess how the model performs on the synthetic dataset; second, we evaluate our method on SVBRDF estimation from real images, when including real-world data, in an unsupervised way, into the training procedure.
A thorough experimental protocol is followed to evaluate the impact of each component of the proposed architecture.
Training and implementation details, as well as additional visual examples (including high-resolution images up to 2048$\times$2048) are reported in the supplementary materials.

\subsection{Datasets and metrics}
\label{sec:datasets}

\subsubsection{Synthetic dataset}
\label{sec:synth_dataset}
We employ the SVBRDF dataset introduced by Deschaintre et al.~\cite{deschaintre2018single}, which is based on the Allegorithmic Substance Share collection\footnote{\url{https://share.substance3d.com/}}.
The entire dataset is made of about 200,000 SVBRDFs, each consisting of a rendered surface with the corresponding diffuse, normal, specular and roughness maps. 

We use the dataset splits provided in~\cite{deschaintre2018single} to train and test our model, as well as a common benchmark to compare our results to the state of the art.
It should be noted that the original rendering of this dataset is performed by assuming a phone-like flash illumination at a fixed distance and centered position. In order to test our model in setups with natural illumination~\cite{li2017modeling}, we generated a variant of the test split of the dataset by re-rendering the same images using random environmental lighting from a library\footnote{\url{https://hdrihaven.com/}}.

\subsubsection{Real-world dataset}
Although synthetic data enables to easily collect large annotated datasets, they tend to diverge from real-world examples, leading to poorly usable models. 
To include natural materials into our training procedure, we collected a large dataset of real-world surfaces, grouped into 14 different categories: \textit{asphalt}, \textit{bark}, \textit{bricks}, \textit{concrete}, \textit{fabric}, \textit{floor}, \textit{foliage}, \textit{granite}, \textit{ground}, \textit{marble}, \textit{metal}, \textit{parquet}, \textit{sand}, \textit{stone}. The dataset consists of several, hand-picked, samples from 3DJungle\footnote{\url{https://3djungle.net/textures/}}, Describable Textures Dataset~\cite{cimpoi14describing} and pictures taken with a smartphone.
The whole dataset consists of 80 images per category, 512$\times$512 each, for a total of 1120 materials. We split the dataset into a training set of 910 samples (65 per class) and a test set of 210 images (15 per class).

\subsection{Model training and evaluation}
We train our networks with mini-batch gradient descent, using the Adam optimizer and a batch size of 6. The learning rate is set to $4\cdot 10^{-5}$ and training is carried out for 250,000 iterations.
The $\alpha$ hyperparameter that weighs the contribution of the adversarial loss is set to 0.2 --- we found that larger values canceled the influence of the reconstruction loss; the $\beta$ hyperparameter, weighing the importance of the MS-SSIM term in the reconstruction loss, is set to 0.84, as suggested in~\cite{zhao2015loss}; we did not find significant differences when modifying this value by small amounts. 

As per common practice in the literature, for quantitative comparison with state-of-the-art methods for single-image SVBRDF estimation, we compute root mean square error (RMSE) between predicted and ground-truth reflectance maps on synthetic images.
We also evaluate RMSE between 5 renderings of the estimated maps, obtained with phone-like flash illumination on different regions of the image (top-left, top-right, bottom-left, bottom-right, center), and the same 5 renderings of the ground truth maps.

Evaluation on real images, due to the lack of a ground-truth set of SVBRDF maps, is carried out qualitatively through visual inspection, by rendering the estimated maps under multiple natural lighting conditions. 

We compare our approach with state-of-the-art methods for single-image SVBRDF estimation. For fair evaluation and to avoid implementation and training biases, we compare with methods that provide source code and pre-trained models, with the same illumination conditions as those on which each model was trained: specifically, we compare with \cite{deschaintre2018single,gao2019deep,deschaintre2019flexible} on single-image SVBRDF estimation with mobile phone flash illumination, and with \cite{li2017modeling} on natural illumination. 

\begin{figure}[ht!]
\centering
\includegraphics[width=1\linewidth]{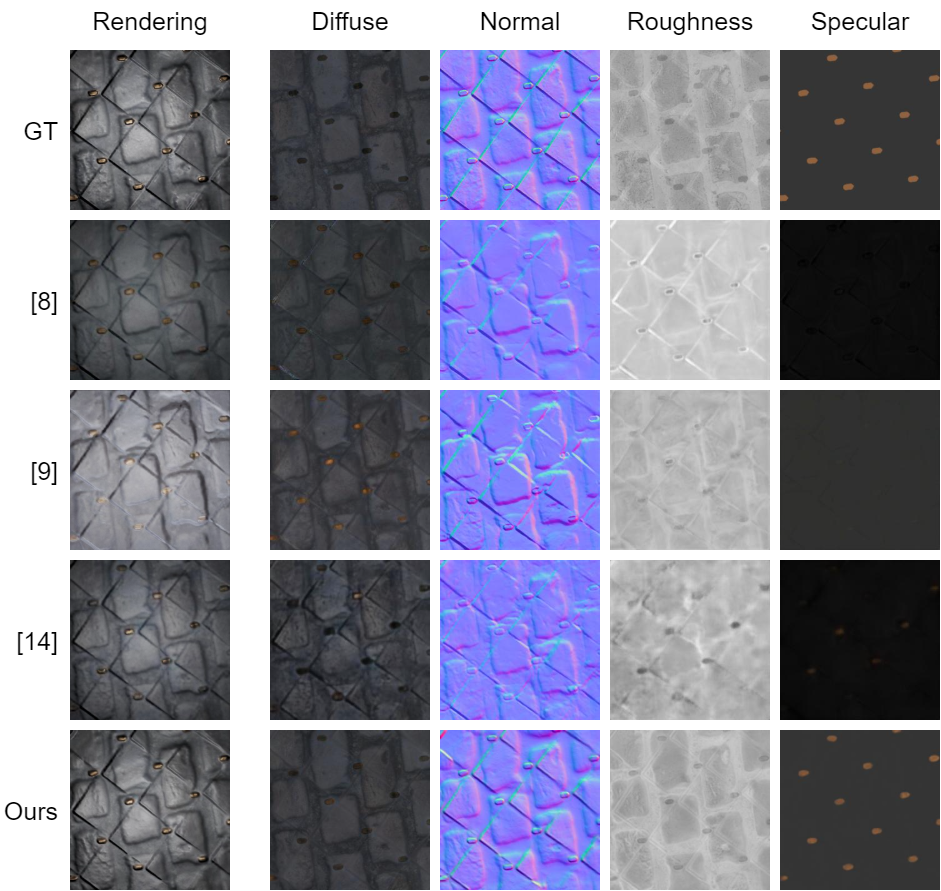}
\caption{\textbf{Qualitative analysis on synthetic images, with flash illumination.} From left to right, the original rendered image and the four SVBRDF maps.}
\label{fig:results_synth_flash}
\end{figure}

\begin{figure}[ht!]
\centering
\includegraphics[width=0.8\linewidth]{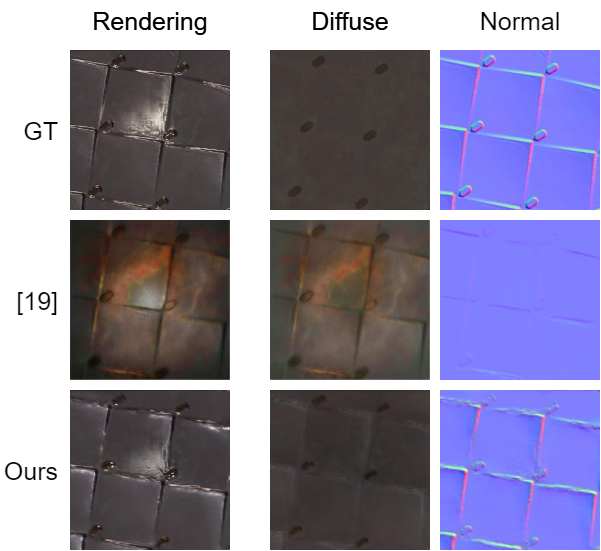}
\caption{\textbf{Qualitative analysis on synthetic images, with natural illumination.} From left to right, the rendered image and the two SVBRDF maps shared by the methods under comparison. Roughness and specular maps are not reported as Li~\cite{li2017modeling} assumes both maps to be homogeneous.}
\label{fig:results_synth_natural}
\end{figure}

\begin{figure*}[ht!]
\centering
\includegraphics[width=0.8\linewidth]{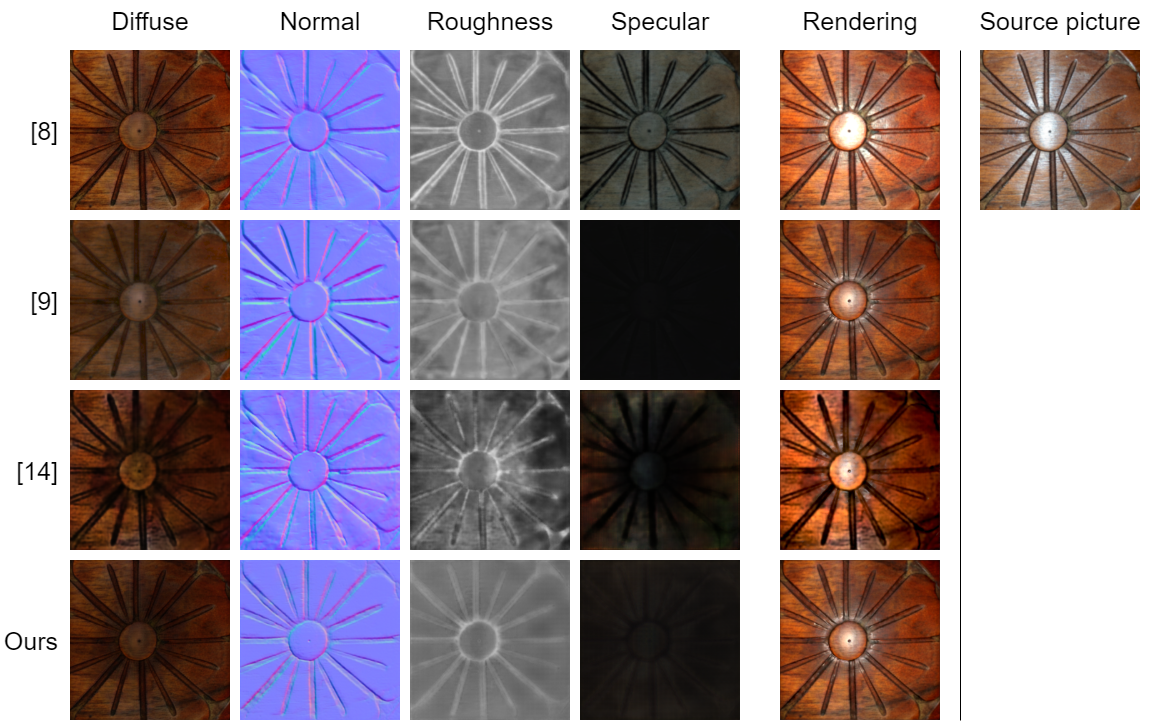}
\caption{\textbf{Qualitative analysis on real images, with flash illumination.} From left to right, the four estimated SVBRDF maps, the corresponding re-rendering and the original image.}
\label{fig:results_real_flash}
\end{figure*}

\subsection{Synthetic acquisition results}
\label{subsec:synth_eval}

In the evaluation with synthetic materials, we report performance obtained by our model when training on synthetic data only.
In Tab.~\ref{tab:results_synth_flash}, we show the results achieved on images illuminated with a mobile phone flash, demonstrating that our approach achieves a higher global accuracy than the methods under analysis.
\begin{table}[h!]
\begin{center}
\begin{tabular}{lrrrrr}
\toprule
\textbf{Method} & \textbf{Diff.} & \textbf{Nrm.} & \textbf{Rgh.} & \textbf{Spec.} & \textbf{Rend.} \\
\midrule
\cite{deschaintre2018single} & 0.019 & 0.035 & 0.129 & 0.050 & 0.083 \\ 
\cite{deschaintre2019flexible} & 0.081 & 0.057 & 0.108 & 0.063 & 0.187 \\ 
\cite{gao2019deep} & 0.050 & 0.062 & 0.119 & 0.202 & 0.108\\ 
\textbf{SurfaceNet} & \textbf{0.017} & \textbf{0.030} & \textbf{0.029} & \textbf{0.014} & \textbf{0.058}\\
\bottomrule
\end{tabular}
\end{center}
\caption{\textbf{Quantitative results on synthetic images, with flash illumination.} Values are reported in terms of RMSE between predicted and ground-truth maps, and between the original image and the corresponding rendering. Column abbreviations correspond to ``Diffuse'', ``Normal'', ``Roughness'', ``Specular'', ``Rendering''. In bold, best results.}
\label{tab:results_synth_flash}
\end{table}

Fig.~\ref{fig:results_synth_flash} presents some examples of estimated SVBRDF maps and the produced rendering of the original image, for better evaluating model prediction. Both our method and~\cite{deschaintre2018single} are able to retrieve normal and roughness maps with satisfactory accuracy and level of detail, but \cite{deschaintre2018single} tends to over-illuminate the diffuse map and lacks fine details in the normal map. \cite{deschaintre2019flexible}, instead, enhances the normal map with more details, but it highlights contrast and flattens the roughness map. 
Finally, the normal maps estimated by~\cite{gao2019deep} are too sensitive to details, over-emphasizing noisy components in the image. Moreover,~\cite{gao2019deep} suffers mainly from a wrong interpretation of the illumination that affects the roughness map, while our method correctly separates light source from material properties. 
Overall, \emph{SurfaceNet} outperforms others methods both quantitatively and qualitatively. 

We then carry out an additional evaluation using synthetic images with natural illumination, as described in Sect.~\ref{sec:datasets}, and compare our method to Li et al.~\cite{li2017modeling}. Tab.~\ref{tab:results_synth_natural} shows the results in terms of RMSE on SVBRDF parameters and rendering. Note that, since Li et al.~\cite{li2017modeling} assume homogenous specular albedo and roughness, we also reduce our corresponding maps to scalar values by computing the average over pixels in each map. On the quantitative comparison, our model outperforms \cite{li2017modeling} on all four parameters. This is also demonstrated by the visual examples reported in Fig.~\ref{fig:results_synth_natural}, showing substantial differences in both the diffuse and normal maps produced by the two methods. In particular, the normal map predicted by Li et al.~\cite{li2017modeling} not only lacks contrast, but also tends to misinterpret light.

\begin{table}
\begin{center}
\begin{tabular}{lrrrrr}
\toprule
\textbf{Method} & \textbf{Diff.} & \textbf{Nrm.} & \textbf{Rgh.} & \textbf{Spec.} & \textbf{Rend.} \\
\midrule
\cite{li2017modeling} & 0.093 & 0.081 & 0.331 & 0.181 & 0.106 \\
SurfaceNet & \textbf{0.033} & \textbf{0.055} & \textbf{0.094} & \textbf{0.041} & \textbf{0.078}\\
\bottomrule
\end{tabular}
\end{center}
\caption{\textbf{Quantitative results on synthetic images, with natural illumination.} Values are reported in terms of RMSE between predicted and ground-truth maps, and between the original image and the corresponding re-rendering. Best results in bold.}
\label{tab:results_synth_natural}
\end{table}

\subsection{Real acquisition results}

In this experiment, we employ both synthetic and real images (the latter used in an unsupervised way) and carry out a qualitative analysis of SVBRDF maps estimated from real-world images (captured with a phone camera and flash illumination).
Fig.~\ref{fig:results_real_flash} shows a qualitative comparison on the estimated maps on real images between our method and \cite{deschaintre2018single,deschaintre2019flexible,gao2019deep}. Similarly to the synthetic evaluation scenario, \emph{SurfaceNet} estimates SVBRDF maps significantly better than existing methods, that, instead, show a few shortcomings: \cite{deschaintre2018single} does not reconstruct  some fine details, especially in the normal map; \cite{deschaintre2019flexible} is able to recover these fine details, but smooths the roughness map and does not correctly estimate the specular map; \cite{gao2019deep} produces a good quality normal map, but over-contrasts the other maps.
Our method is, instead, able to generate a smoother roughness map, while preserving edges and fine details.

\begin{figure}[ht!]
\centering
\includegraphics[width=1\linewidth]{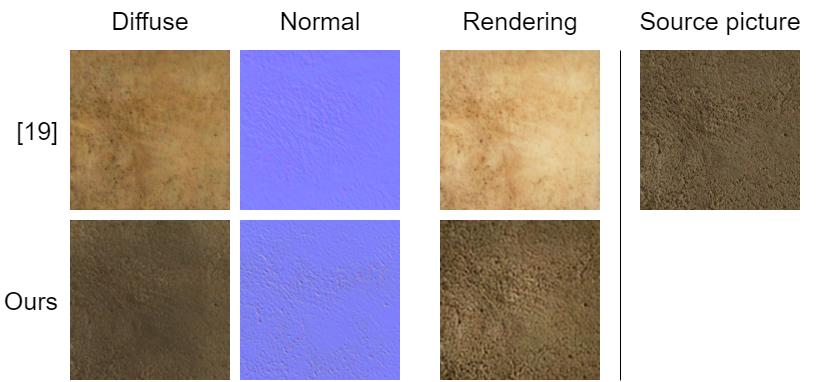}
\caption{\textbf{Qualitative analysis on real images, with natural illumination.} From left to right, the two estimated SVBRDF maps shared by the methods under comparison, the corresponding re-rendering and the original image.}
\label{fig:results_real_natural}
\end{figure}

Finally, we again compare our approach to Li et al~\cite{li2017modeling} on real images with natural illumination.  Fig.~\ref{fig:results_real_natural} shows that our proposed method is able to capture diffuse and normal information remarkably well, as demonstrated by the rendered image resembling the source image more than~\cite{li2017modeling}.  

\begin{figure}
    \centering
    \includegraphics[width=0.48\textwidth]{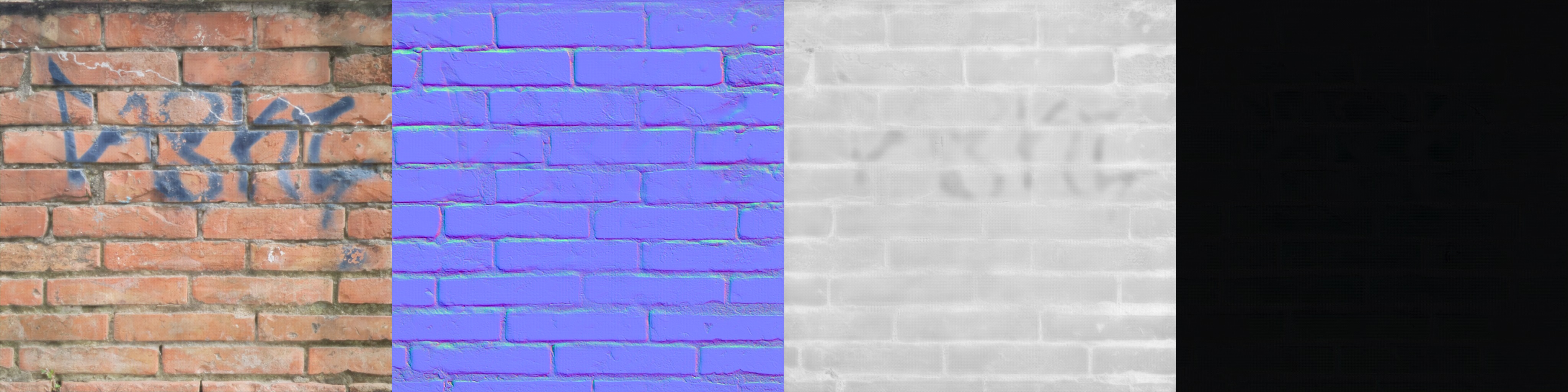}
    \caption{High-resolution (1024$\times$1024) SVBRDF estimation on real-world materials (zoom in to see fine details). The picture was taken with a smartphone camera under natural illumination.}
    \label{fig:high_res}
\end{figure}

\begin{figure}
    \centering
    \includegraphics[width=0.48\textwidth]{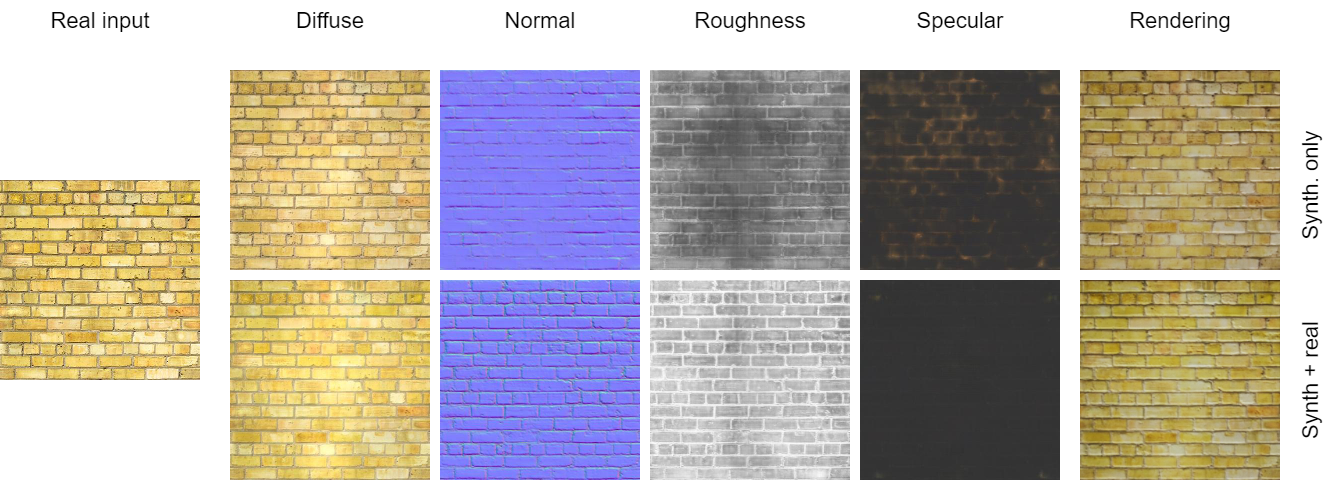}
    \caption{Qualitative comparison of SurfaceNet SVBRDF estimation for real data when training only with synthetic data (first row) and  (second row) with real data. Zoom-in to see details.}
    \label{fig:domain_adapt}
\end{figure}

Furthermore, one of the major strengths of our approach is its ability to scale up to high resolutions with good estimation quality. 
A qualitative example of SVBRDF estimation on a real-world 1024$\times$1024 images is given in Fig.~\ref{fig:high_res}.

\subsection{Ablation study}

We perform an ablation study, on the synthetic image dataset, to substantiate our architectural design and training strategy choices. 
We first perform some control experiments to substantiate the SurfaceNet architecture. As a baseline, we use DeepLab-v3 encoder and interpolate features to output resolution; we then add (in order) a decoder network with learnable upsampling layers, and skip connections with downsampled input.
We then evaluate the impact of a patch-based discriminator w.r.t.  a standard image-based discriminator trained adversarially using the full training loss described in Sect.~\ref{sec:training}. Results in Tab.~\ref{tab:ablation_arch} show that all the architectural changes to the baseline positively affect the accuracy of estimation. However, the highest contribution to the final performance is given when adding the adversarial training procedure at the patch level.

We finally evaluate the importance of the different loss terms employed during training. Tab.~\ref{tab:ablation_loss} shows the estimation accuracy when using a supervised loss alone (simply $L_1$ first, and then adding MS-SSIM to obtain $\mathcal{L}_\text{sup}$), an unsupervised (i.e., adversarial) loss alone ($\mathcal{L}_\text{unsup}$), and the overall $\mathcal{L}$ loss, both when only synthetic images are used ($\mathcal{L}$ ``synth'') and when real data are included ($\mathcal{L}$ ``real''). 
Results provided in Tab.~\ref{tab:ablation_arch} confirm that all the inclusion of the adversarial loss significantly improves our method's accuracy. It is interesting to note that, when real images are included in the training with the full loss ($\mathcal{L}$ ``real''), performance slightly degrades w.r.t. using synthetic images only. This is expected, since including real images during training acts as a regularizer that prevents the model from learning features that are specific to synthetic images. However, this regularization effect results in improved accuracy when testing our model on real images: the example in Fig.~\ref{fig:domain_adapt} shows that, in this case, training on real images significantly improves SVBRDF estimation.

\begin{table}[h!]
\begin{center}
\begin{tabular}{llrrrrr}
\toprule
& & \textbf{Diff.} & \textbf{Nrm.} & \textbf{Rgh.} & \textbf{Spec.} & \textbf{Rend.} \\
\midrule
\parbox[t]{2mm}{\multirow{4}{*}{\rotatebox[origin=c]{90}{Gen.}}} & Base & 0.207 & 0.156 & 0.322 & 0.138 & 0.185 \\
& + Dec. & 0.086 & 0.093 & 0.130 & 0.042 & 0.087 \\
& \hspace{0.15cm} + Skip & 0.073 & 0.122 & 0.253 & 0.084 & 0.073 \\
\midrule
\parbox[t]{2mm}{\multirow{2}{*}{\rotatebox[origin=c]{90}{Disc.}}} 
& 
\hspace{0.35cm} + Image & 0.021 & 0.037 & 0.065 & 0.022 & 0.067 \\
& 
\hspace{0.35cm} + \textbf{Patch} & \textbf{0.017} & \textbf{0.030} & \textbf{0.029} & \textbf{0.014} & \textbf{0.058} \\
\bottomrule
\end{tabular}
\end{center}
\caption{\textbf{Ablation study of design choices.} We ablate the generator by adding, to a ``base'' variant of our model (no skip connections, upsampling through interpolation, no adversarial loss) the following blocks: 1) learnable upsampling layers (``Decoder''); 2) skip connections with downsampled input (``Skip''); 3) discriminator network and adversarial loss (``Full''). We also ablate the discriminator network, when carrying out patch-based or image-based discrimination.}
\label{tab:ablation_arch}
\end{table}

\begin{table}[h!]
\begin{center}
\begin{tabular}{lrrrrr}
\toprule
\textbf{Losses} & \textbf{Diff.} & \textbf{Nrm.} & \textbf{Rgh.} & \textbf{Spec.} & \textbf{Rend.} \\
\midrule
$L_1$ & 0.162 & 0.147 & 0.335 & 0.184 & 0.134 \\
$\mathcal{L}_{\text{sup}}$ & 0.073 & 0.122 & 0.253 & 0.084 & 0.088 \\
$\mathcal{L}_{\text{unsup}}$ & 0.133  & 0.112 & 0.274  & 0.119 & 0.128 \\
$\mathcal{L}$ (synth) & 0.017 & 0.030 & 0.029 & 0.014 & 0.058 \\
$\mathcal{L}$ (real) & 0.022 & 0.033 & 0.043 & 0.022 & 0.064 \\

\bottomrule
\end{tabular}
\end{center}
\caption{\textbf{Ablation study of used losses.} We evaluate the performance of our method when using a supervised loss only ($L_1$ and $\mathcal{L}_\text{sup}$), an unsupervised loss only ($\mathcal{L}_\text{unsup}$), and our full loss ($\mathcal{L}$), when training on synthetic data only (``synth'') and when including real images (``real'').}
\label{tab:ablation_loss}
\end{table}

\section{Conclusion}
In this paper, we present an adversarial learning-based approach, \textit{SurfaceNet}, for the estimation of SVBRDF material reflectance parameters from single images.
The proposed approach specifically leverages generative adversarial networks (GANs) to reconstruct high-quality, high-resolution SVBRDF maps as well as to enable unsupervised prediction on real-world samples (for which no ground-truth annotations are available) by forcing the model to learn domain-independent features that are applicable to both synthetic and real data. 
Experimental results show that our model quantitatively and qualitatively outperforms state-of-the-art methods. More interestingly, it shows that the inclusion of real samples through the proposed unsupervised training procedure significantly impacts the quality of reflectance parameter estimation for real images. Thus, the proposed adversarial strategy demonstrates a remarkable ability in filling the gap between synthetic and real data distributions,  allowing deep models to perform well with real-world images, despite being trained supervisedly with synthetic data only.


{\small
\begin{thebibliography}{10}\itemsep=-1pt

\bibitem{aittala2016reflectance}
Miika Aittala, Timo Aila, and Jaakko Lehtinen.
\newblock Reflectance modeling by neural texture synthesis.
\newblock {\em ACM Transactions on Graphics (ToG)}, 35(4):1--13, 2016.

\bibitem{aittala2013practical}
Miika Aittala, Tim Weyrich, and Jaakko Lehtinen.
\newblock Practical svbrdf capture in the frequency domain.
\newblock {\em ACM Trans. Graph.}, 32(4):110--1, 2013.

\bibitem{bi2020deep}
Sai Bi, Zexiang Xu, Kalyan Sunkavalli, David Kriegman, and Ravi Ramamoorthi.
\newblock Deep 3d capture: Geometry and reflectance from sparse multi-view
  images.
\newblock In {\em Proceedings of the IEEE/CVF Conference on Computer Vision and
  Pattern Recognition}, pages 5960--5969, 2020.

\bibitem{chen2014reflectance}
Guojun Chen, Yue Dong, Pieter Peers, Jiawan Zhang, and Xin Tong.
\newblock Reflectance scanning: estimating shading frame and brdf with
  generalized linear light sources.
\newblock {\em ACM Transactions on Graphics (TOG)}, 33(4):1--11, 2014.

\bibitem{chen2017deeplab}
Liang-Chieh Chen, George Papandreou, Iasonas Kokkinos, Kevin Murphy, and Alan~L
  Yuille.
\newblock Deeplab: Semantic image segmentation with deep convolutional nets,
  atrous convolution, and fully connected crfs.
\newblock {\em IEEE transactions on pattern analysis and machine intelligence},
  40(4):834--848, 2017.

\bibitem{chen2017rethinking}
Liang-Chieh Chen, George Papandreou, Florian Schroff, and Hartwig Adam.
\newblock Rethinking atrous convolution for semantic image segmentation.
\newblock {\em arXiv preprint arXiv:1706.05587}, 2017.

\bibitem{cimpoi14describing}
M. Cimpoi, S. Maji, I. Kokkinos, S. Mohamed, , and A. Vedaldi.
\newblock Describing textures in the wild.
\newblock In {\em Proceedings of the {IEEE} Conf. on Computer Vision and
  Pattern Recognition ({CVPR})}, 2014.

\bibitem{deschaintre2018single}
Valentin Deschaintre, Miika Aittala, Fredo Durand, George Drettakis, and Adrien
  Bousseau.
\newblock Single-image svbrdf capture with a rendering-aware deep network.
\newblock {\em ACM Transactions on Graphics (ToG)}, 37(4):1--15, 2018.

\bibitem{deschaintre2019flexible}
Valentin Deschaintre, Miika Aittala, Fr{\'e}do Durand, George Drettakis, and
  Adrien Bousseau.
\newblock Flexible {SVBRDF} capture with a multi-image deep network.
\newblock In {\em Computer Graphics Forum}, volume~38, pages 1--13. Wiley
  Online Library, 2019.

\bibitem{deschaintre2020guided}
Valentin Deschaintre, George Drettakis, and Adrien Bousseau.
\newblock Guided fine-tuning for large-scale material transfer.
\newblock In {\em Computer Graphics Forum}, volume~39, pages 91--105. Wiley
  Online Library, 2020.

\bibitem{dong2019deep}
Yue Dong.
\newblock Deep appearance modeling: A survey.
\newblock {\em Visual Informatics}, 3(2):59--68, 2019.

\bibitem{dorsey2005digital}
Julie Dorsey and Holly Rushmeier.
\newblock Digital modeling of the appearance of materials.
\newblock In {\em ACM SIGGRAPH 2005 Courses}, pages 1--es. 2005.

\bibitem{10.5555/1557600}
Julie Dorsey, Holly Rushmeier, and Franois Sillion.
\newblock {\em Digital Modeling of Material Appearance}.
\newblock Morgan Kaufmann Publishers Inc., San Francisco, CA, USA, 2007.

\bibitem{gao2019deep}
Duan Gao, Xiao Li, Yue Dong, Pieter Peers, Kun Xu, and Xin Tong.
\newblock Deep inverse rendering for high-resolution svbrdf estimation from an
  arbitrary number of images.
\newblock {\em ACM Trans. Graph.}, 38(4):134--1, 2019.

\bibitem{guo2020materialgan}
Yu Guo, Cameron Smith, Milo{\v{s}} Ha{\v{s}}an, Kalyan Sunkavalli, and Shuang
  Zhao.
\newblock Materialgan: reflectance capture using a generative svbrdf model.
\newblock {\em arXiv preprint arXiv:2010.00114}, 2020.

\bibitem{he2016deep}
Kaiming He, Xiangyu Zhang, Shaoqing Ren, and Jian Sun.
\newblock Deep residual learning for image recognition.
\newblock In {\em Proceedings of the IEEE conference on computer vision and
  pattern recognition}, pages 770--778, 2016.

\bibitem{isola2017image}
Phillip Isola, Jun-Yan Zhu, Tinghui Zhou, and Alexei~A Efros.
\newblock Image-to-image translation with conditional adversarial networks.
\newblock In {\em Proceedings of the IEEE conference on computer vision and
  pattern recognition}, pages 1125--1134, 2017.

\bibitem{larsen2016autoencoding}
Anders Boesen~Lindbo Larsen, S{\o}ren~Kaae S{\o}nderby, Hugo Larochelle, and
  Ole Winther.
\newblock Autoencoding beyond pixels using a learned similarity metric.
\newblock In {\em International conference on machine learning}, pages
  1558--1566. PMLR, 2016.

\bibitem{li2017modeling}
Xiao Li, Yue Dong, Pieter Peers, and Xin Tong.
\newblock Modeling surface appearance from a single photograph using
  self-augmented convolutional neural networks.
\newblock {\em ACM Transactions on Graphics (ToG)}, 36(4):1--11, 2017.

\bibitem{li2018materials}
Zhengqin Li, Kalyan Sunkavalli, and Manmohan Chandraker.
\newblock Materials for masses: Svbrdf acquisition with a single mobile phone
  image.
\newblock In {\em Proceedings of the European Conference on Computer Vision
  (ECCV)}, pages 72--87, 2018.

\bibitem{li2018learning}
Zhengqin Li, Zexiang Xu, Ravi Ramamoorthi, Kalyan Sunkavalli, and Manmohan
  Chandraker.
\newblock Learning to reconstruct shape and spatially-varying reflectance from
  a single image.
\newblock {\em ACM Transactions on Graphics (TOG)}, 37(6):1--11, 2018.

\bibitem{lombardi2012single}
Stephen Lombardi and Ko Nishino.
\newblock Single image multimaterial estimation.
\newblock In {\em 2012 IEEE Conference on Computer Vision and Pattern
  Recognition}, pages 238--245. IEEE, 2012.

\bibitem{walter2007microfacet}
Bruce Walter, Stephen~R Marschner, Hongsong Li, and Kenneth~E Torrance.
\newblock Microfacet models for refraction through rough surfaces.
\newblock {\em Rendering techniques}, 2007:18th, 2007.

\bibitem{wang2003multiscale}
Zhou Wang, Eero~P Simoncelli, and Alan~C Bovik.
\newblock Multiscale structural similarity for image quality assessment.
\newblock In {\em The Thrity-Seventh Asilomar Conference on Signals, Systems \&
  Computers, 2003}, volume~2, pages 1398--1402. Ieee, 2003.

\bibitem{zhao2015loss}
Hang Zhao, Orazio Gallo, Iuri Frosio, and Jan Kautz.
\newblock Loss functions for neural networks for image processing.
\newblock {\em arXiv preprint arXiv:1511.08861}, 2015.

\bibitem{zhou2016sparse}
Zhiming Zhou, Guojun Chen, Yue Dong, David Wipf, Yong Yu, John Snyder, and Xin
  Tong.
\newblock Sparse-as-possible svbrdf acquisition.
\newblock {\em ACM Transactions on Graphics (TOG)}, 35(6):1--12, 2016.

\end{thebibliography}
}
\end{document}